%% file: main.tex
\pdfoutput=1
\documentclass[11pt]{article}

\usepackage{authblk}
\usepackage{ACL2023}
\usepackage{times}
\usepackage{latexsym}
\usepackage[T1]{fontenc}
\usepackage{booktabs}
\usepackage{amsmath}
\usepackage[utf8]{inputenc}
\usepackage{microtype}
\usepackage{inconsolata}
\usepackage{fancyhdr}
\usepackage{graphicx}
\usepackage{caption}
\usepackage{subcaption}
\usepackage{tcolorbox}
\usepackage{hyperref}
\usepackage{amssymb}

\newcommand{\concept}[1]{\noindent\textbf{#1}\quad}
\newcommand{\model}{LLaMa-SciQ}
\newcommand{\titlelong}{An Educational Chatbot for Answering Science MCQ}

\newcommand{\generation}[4]{
    \textbox{
        #1\\

        \texttt{\textbf{\#\#\#Question:}}\\
        #2\\

        \texttt{\textbf{\#\#\#Answer:}}\\
        #3
    }{#4}
}

\newcommand{\textbox}[2]{
    \begin{tcolorbox}[colback=blue!5, colframe=black!75, width=\columnwidth, boxrule=0.2mm, arc=2mm, auto outer arc, title={#2}]
    #1
    \end{tcolorbox}
}

\newcommand{\sample}[2]{
    \textbox{
        \texttt{\{}\\
        #1\\
        \texttt{\}}
    }{Sample from the #2}
}

\def\titleline{\rule{\textwidth}{0.4mm}}

\title{\titleline\\ \vspace{12pt} 
        \model: \\ \titlelong
\\\titleline}

\author[1]{\textbf{Marc-Antoine Allard}}
\author[1]{\textbf{Matin Ansaripour}}
\author[1]{\textbf{Maria Yuffa}}
\author[1]{\textbf{Paul Teiletche}}
\affil[1]{EPFL, Lausanne, The Wordsmiths}
\affil[ ]{\texttt{\{firstname.lastname\}@epfl.ch}}

\begin{document}
\maketitle
\thispagestyle{fancy}
\thispagestyle{empty}

\begin{abstract}
Large Language Models (LLMs) often struggle with tasks requiring mathematical reasoning, particularly multiple-choice questions (MCQs). To address this issue, we developed \model{}, an educational chatbot designed to assist college students in solving and understanding MCQs in STEM fields. We begin by fine-tuning and aligning the models to human preferences. After comparing the performance of Mistral-7B and LLaMa-8B, we selected the latter as the base model due to its higher evaluation accuracy. To further enhance accuracy, we implement Retrieval-Augmented Generation (RAG) and apply quantization to compress the model, reducing inference time and increasing accessibility for students. For mathematical reasoning, \model{} achieved 74.5\% accuracy on the GSM8k dataset and 30\% on the MATH dataset. However, RAG does not improve performance and even reduces it, likely due to retriever issues or the model's unfamiliarity with context. Despite this, the quantized model shows only a 5\% loss in performance, demonstrating significant efficiency improvements.


\end{abstract}

\section{Introduction}
\label{sec:intro}
Large Language Models (LLMs) are known to perform poorly on questions requiring advanced mathematical reasoning \cite{wu2023empirical}. This is especially true for the university level problems \cite{wang2024scibench}. In literature, the failure of current approaches is attributed to inability of LLMs to recognize and correct a wrong answer \cite{imani2023mathprompter} as well as catastrophic forgetting of linguistic skills when trained on maths data \cite{sharma2023learning}. The issues cannot be fully addressed with a simple prompting strategy due to data variability \cite{wang2024scibench}. 

This project explores state-of-the-art LLMs for creation of an accessible chatbot that assists students in mathematics, physics and computer science. Specifically, we fine-tune LLaMa-3-8B model as well as Mistral-7B on a variety of mathematical and scientific datasets further using Direct Preference Optimization (DPO) to align model's responses to the ones preferred by a student. We compare the performances of the models and demonstrate the significantly superior performance of the LLaMa model with which we proceed. 
We try to enhance the accuracy of fine-tuned LLaMa-3-8B model by applying Retrieval Augmented Generation (RAG). Finally, we quantize the LLM for more efficient inference, making it suitable for students needs.

\section{Related Work}
\label{sec:rw}
With recent release of ChatGPT-3.5 and ChatGPT-4, the number of people using LLMs for education has sky-rocketed \cite{futterer2023chatgpt}. To leverage its capabilities while making user-friendly interfaces vast amount of research is dedicated to creation of LLM based Chatbots for academic purposes \cite{10.1145/3627508.3638293}. Despite success of ChatGPT models on lingustic tasks, their performance was limited on problems involving mathematical reasoning. This is especially true for MCQ questions where the answer is not verbal. This is showcased by the work of \cite{savelka2023large}, where GPT model struggled to give the correct answer to the questions that do not contain natural language.

To improve the performance of the pre-trained LLM model on mathematical questions while ensuring the alignment of the responses with the intended purposes and human values, we considered both Supervised Fine-tuning on mathematical and scientific datasets as well as DPO on the preference pairs ranked by students. This approach was inspired by InstructGPT \cite{ouyang2022training} in aligning LLMs with human preferences. We also considered DPO with an offset \cite{amini2024direct}. This approach introdues variability in treatment of preference pairs and could be less robust, since the inferred offset value might be high in a mis-annotated responses and confuse the model. Therefore, to achieve good results on noisy data while keeping the implementation simple, we chose Conservative DPO (cDPO) loss for DPO fine-tuning strategy, primarily due to its robustness on noisy data.

Further refining the model, we have considered RAG. Initially, we aimed to use pre-trained RAG retriever \cite{karpukhin-etal-2020-dense} or adopting the novel concept of Retrieval Augmented Fine-Tuning (RAFT) \cite{zhang2024raft}. Thoroughly reviewing the literature \cite{gao2024retrievalaugmented} we yielded to the \textit{Naive RAG} strategy due to good performance and straightforwardness of the approach. 

Finally, we considered quantizing the model to reduce the computational costs when using the Chatbot while maintaining good response accuracy. At first, we sought to use QuIP \cite{chee2024quip} and recently released QuIP\# \cite{tseng2024quip} due to its ability to leverage incoherence in weights and Hessian matrices. We have also considered QLoRA \cite{hu2021lora}, which is not computationally demanding, yet preserves the 16-bit fine-tuning task performance of LLMs. Before attempting the advanced methods, we have tried GPTQ \cite{frantar2023gptq}. Nonetheless, facing some numerical issues with quantizing our model with GPTQ, we decided to use the 4-bit quantization provided by Unsloth bitsandbytes library.

Overall, our work is a nice step towards creating an efficient, student-oriented educational assistant for questions requiring mathematical reasoning.

\section{Approach}
\label{sec:approach}
Our approach consisted of performing SFT training on both Mistral-7B and LLaMa-3-8B. We then compared the performance of two models with SFT and DPO training and proceeded with LLaMa-3-8B which performed better on the evaluation set (Figure \ref{tab:results-mistral-llama}). In this section, we outline the details of the models and their fine-tuning strategy with emphasis on LLaMa-3-8B.
\subsection{Base Model Architecture}
\concept{LLaMa-3-8B} \cite{llama3modelcard} is an auto-regressive language model featuring an enhanced transformer architecture with a standard decoder-only design. The model integrates supervised fine-tuning (SFT) and reinforcement learning with human feedback (RLHF) to better align with human preferences regarding safety and helpfulness. Llama 3, which uses a tokenizer with a 128K-token vocabulary for more efficient language encoding, shows significant performance improvements over its predecessor. The model, trained on sequences up to 8,192 tokens with boundary-aware self-attention, uses Grouped-Query Attention (GQA) to enhance inference scalability.\\ 

\concept{Mistral-7B} \cite{jiang2023mistral}, a language model with 7 billion parameters, utilizes a transformer-based architecture comprising multiple transformer blocks. It employs sliding window attention that allows the model to attend to tokens outside of the window, Rolling Buffer Cache to reduce the cache memory usage while keeping the model quality. It also utilizes pre-fill and chunking, which involve loading known parts of a prompt into the (k, v) cache to facilitate token generation. If the prompt is lengthy, it is segmented into smaller chunks, each pre-filled into the cache to enhance processing efficiency during token prediction.

\subsection{Training Pipeline}
\begin{figure*}[h]
    \centering
        \includegraphics[width=\textwidth]{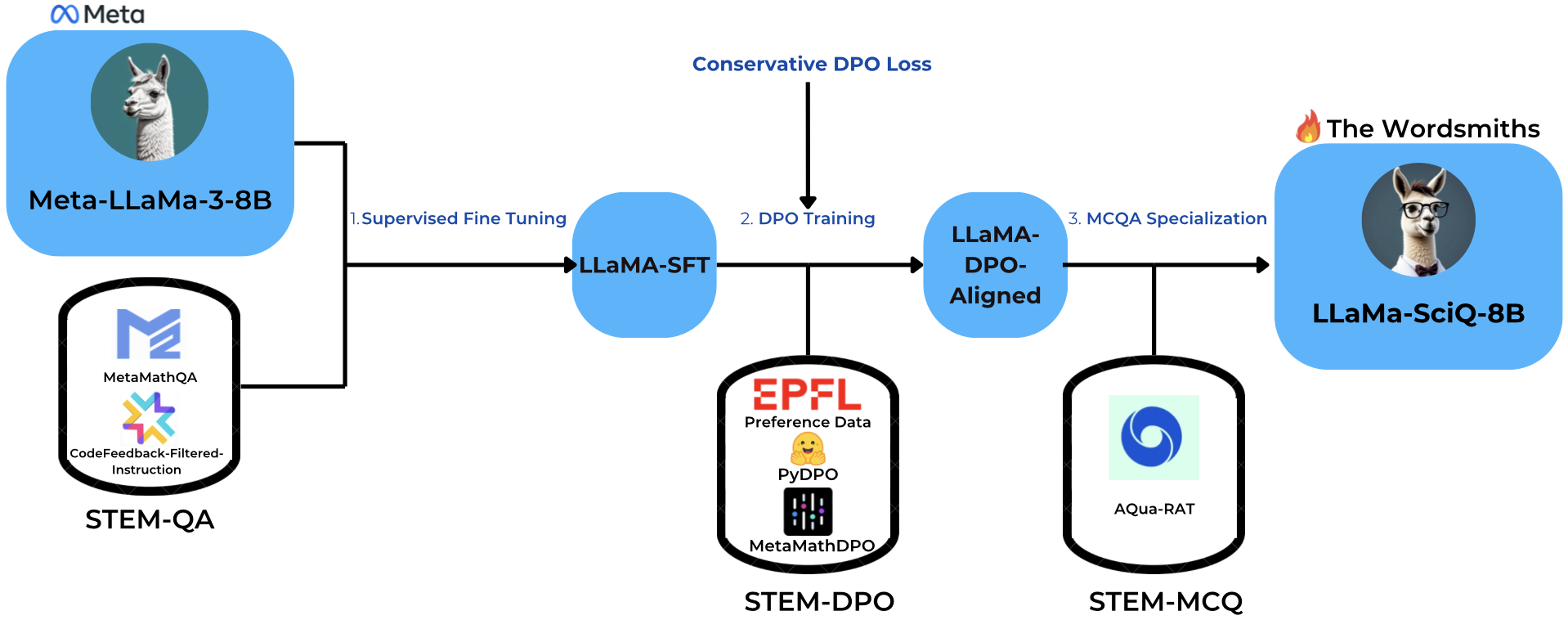}
        \caption{The Training Pipeline: Organized into three consecutive stages; Supervised Fine-Tuning, Direct Preference Optimization Training, and Multiple Choice Question Answering Specialization.}
        \label{fig:pipeline}
\end{figure*}

The training pipeline for LLaMa-3-8B model is demonstrated in Figure \ref{fig:pipeline}. We first performed Supervised Fine-tuning on a mix of specialized maths and science datasets. We then performed DPO training using preference data generated and annotated by students via cDPO loss. Finally, we gauged the performance of the model on \texttt{AQuA-Rat}~\cite{ling2017program} dataset which contains STEM-related MCQ questions.

Mistral-7B used the same process as LLaMa, except for the final SFT, since LLaMa showed superior performance (Figure \ref{tab:results-mistral-llama}). Ultimately, we have not implemented both due to time constraints.

\subsubsection{Supervised Fine Tuning}
The results of supervised fine-tuning of two models are demonstrated in Table \ref{tab:results-mistral-llama}.
\subsubsection{Preference Data Collection}
To collect preference data, a cohort of 300 students was asked to generate two responses, a better one and a slightly worse one but preferably still correct, to the question using GPT-wrapper. The students were further asked to rank the responses.

To generate answers for the dataset of questions in mathematics, physics, and computer science, we have developed a prompting strategy that incorporates several techniques. Firstly, we create a separate chat for each subject id. Secondly, we use Chain-of-Thought (CoT)~\cite{wei2022chain}, which guides the model to reach conclusions in a step-by-step manner. Thirdly, the model is prompted with the instruction provided in \ref{appendix:dpo_examples_instruct}. Finally, for generating preference pairs, we employ the following method: to achieve a better response, we prompt the model to re-read the question before attempting to solve it. This methods was shown by \cite{xu2024rereading} to consistently improve performance for LLMs, except for Vanilla ChatGPT. For the worst answer, the model is instructed to provide a very brief explanation.

\subsubsection{Reward Model}
The reward model is a critical component of the DPO fine-tuning strategy. The reward model is based on the policy that maximizes the reward with KL constraint to the reference policy:
\[
\pi^* = \arg\max_{\pi} \mathbb{E}_{y \sim \pi} \left[ r(y) - \beta \log \frac{\pi(y)}{\pi_{\text{ref}}(y)} \right]
\]
Considering a small probability that the preference pair could be flipped, the preferred response is in reality less correct or explicit than the other one, we can derive the following DPO loss to optimise the reward model:
\begin{align}
    \mathcal{L}^{\epsilon}_{DPO}(\theta, y_w, y_l) &= -(1 - \epsilon) \log \hat{p}_{\theta}(y_w > y_l) \\
    &\quad - \epsilon \log (1 - \hat{p}_{\theta}(y_w > y_l)) \\
    &= (1 - \epsilon) \mathcal{L}_{DPO}(\theta, y_w, y_l) \\
    &+ \epsilon \mathcal{L}_{DPO}(\theta, y_l, y_w),
\end{align}
where $\epsilon$ indicates the probability of the answer being wrong (or flipping the pair).

\begin{table}[h!]
\centering
\begin{tabular}{lll}
\toprule
\textbf{Base Model} & \textbf{Strategy} & \textbf{Test Rwrds Acc.}  \\
\midrule
LLaMa-3-8B & DPO & \textbf{79.7\%} \\
LLaMa-3-8B & SFT+DPO & 79.3\% \\
Mistral-7B & DPO & 76.5\% \\
Mistral-7B & SFT+DPO & 71.3\% \\
\bottomrule
\end{tabular}
\caption{Accuracy scores of the models on 1000 samples of the test set.}
\label{tab:results-mistral-llama}
\end{table}

\subsection{Retrieval Augmented Generation}
\begin{figure}[h]
    \centering
        \includegraphics[width=0.55\columnwidth, height=1.2\columnwidth]{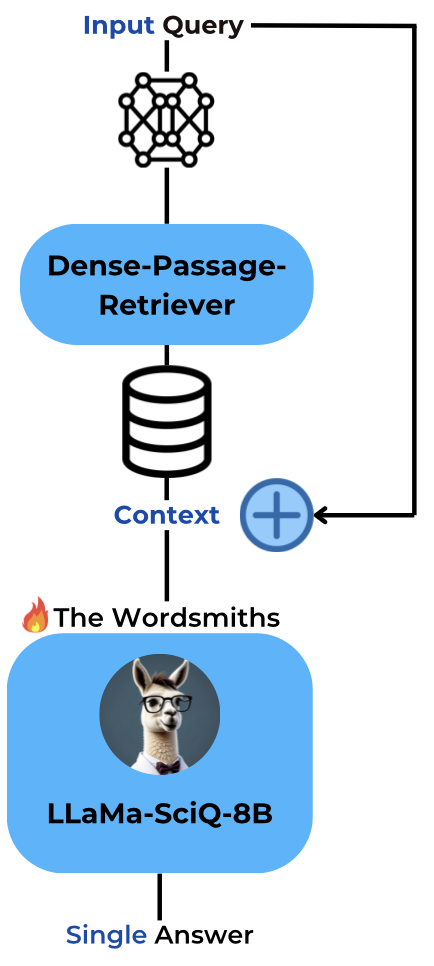}
        \caption{The RAG Pipeline}
        \label{fig:rag}
\end{figure}

We augment \model{} by incorporating Retrieval-Augmented Generation (RAG) methods. This approach stands out as one of the most effective means to enhance the predictive capabilities of our model. RAG combines the capabilities of generative models, dense vector indices of a racorpus of documents, and pre-trained neural retrievers. 
Figure~\ref{fig:rag} summarizes our RAG pipeline, known as the \texttt{Naive RAG}.
We use the dataset described in~\ref{sec:dpr_dataset} as our Dense Passage Retrieval (DPR) corpus of documents over which we create an index using Facebook's FAISS library~\cite{douze2024faiss}. Documents are then retrieved using Facebook's DPR question encoder~\cite{karpukhin-etal-2020-dense} and added to the prompt in the format detailed in Appendix~\ref{appendix:prompt}.

We observed that our model is getting biased by saying to use the provided information. So we changed the prompt for RAG to tell the model to consider the model but not get biased on the information and try to fulfill the objective of the questions.

\subsection{Quantization}

We took an alternative root compared to standard quantization techniques. In particular, we specified the bytes-and-bits (bnb) parameter when loading the model using Unlosth package. We, therefore, reduced the weights to 4-bits while sustaining the accuracy.

When you enable "load\_in\_4bits" in the "from\_pretrained" function of the unsloth repository, the model utilizes a quantization technique facilitated by the bitsandbytes library. This technique allows the model weights to be represented with 4-bit precision, significantly reducing the model's memory footprint while attempting to preserve performance.

This 4-bit quantization primarily involves the transformation of model weights, previously in full-precision formats like fp16 or bf16, into a 4-bit format. The process entails creating instances of linear layers designed for 4-bit operations (e.g., "Linear4bit"), and then loading the original model's weights into these quantized modules. The actual quantization happens when these modified models are transferred to a computation device like a GPU.

This quantization approach can utilize different data types for quantization, like FP4 (Float4) or NF4 (NormalFloat4), which are tailored for different kinds of data distributions and usage scenarios. For example, NF4 is designed for data that naturally follows a normal distribution, offering potential performance improvements in such cases.

\section{Experiments}
\subsection{Data}
\label{sec:data}
This section outlines the datasets created for our model's alignment stages. Samples of the datasets can be found in Appendix~\ref{appendix:dataset_examples}.

\subsubsection{SFT Dataset}
\label{sec:sft_dataset}
We first introduce \texttt{StemQA}, a specialized dataset to extend our model's performance on math and coding questions. This dataset is a blend of \texttt{MetaMathQA}~\cite{yu2023metamath} and \texttt{CodeFeedback-Filtered-Instruction}~\cite{zheng2024opencodeinterpreter} datasets. It is balanced so that 75\% of the questions are math-related, while the remaining 25\% are coding-related. Table~\ref{tab:sft_dataset_sizes} presents these proportions. The answers now include the rationale followed by \textit{"The answer is: <Maths/Code>"} to simplify future answer extraction.\\

\begin{table}[h]
    \centering
    \begin{tabular}{l c c}
        \toprule
        \textbf{Dataset} & \textbf{Size} & \textbf{Ratio} \\
        \midrule
        \texttt{MetaMathQA} & 375,000 & 75\% \\
        \texttt{CodeFeedback} & 125,000 & 25\% \\
        \midrule
        \textbf{\texttt{StemQA} (ours)} & 500,000 & -- \\
        \bottomrule
    \end{tabular}
    \caption{Dataset sizes and their ratios of the SFT dataset}
    \label{tab:sft_dataset_sizes}
\end{table}

\concept{MetaMathQA}
Augmented version of the \underline{training} sets from \texttt{GSM8K}~\cite{cobbe2021training} and \texttt{MATH}~\cite{hendrycks2021measuring}. \\

\concept{CodeFeedback-Filtered-Instruction} 
Curated collection of code instruction queries extracted from four prominent open-source code instruction tuning datasets.

\subsubsection{DPO Dataset}
\label{sec:dpo_dataset}
Then, we introduce \texttt{StemDPO}, a dataset to align our model with human preferences, focusing particularly on STEM questions. This dataset combines our class preference pairs with the \texttt{PyDPO} and \texttt{MetaMathDPO} datasets. Our objective was to expand this dataset to a size of 50,000 samples, maintaining the same distribution proportions as the SFT dataset, assuming our class preferences are similarly balanced (see Table~\ref{tab:dpo_dataset_sizes}).\\

\begin{table}[h]
    \centering
    \begin{tabular}{l c c}
        \toprule
        \textbf{Dataset} & \textbf{Size} & \textbf{Ratio} \\
        \midrule
        \texttt{ClassPreferences} & 21,596 & 43\% \\
        \texttt{PyDPO} & 7,101 & 14\% \\
        \texttt{MetaMathDPO} & 21,303 & 43\% \\
        \midrule
        \textbf{\texttt{StemDPO} (ours)} & 50,000 & -- \\
        \bottomrule
    \end{tabular}
    \caption{Dataset sizes and their ratios of the DPO dataset}
    \label{tab:dpo_dataset_sizes}
\end{table}

\concept{PyDPO} 
DPO dataset meant to enhance python coding abilities. This dataset uses the excellent \texttt{Tested-22k-Python-Alpaca} dataset as the "chosen" responses and generates the "rejected" values with a mix of \texttt{airoboros-l2-13b-3.1} and \texttt{bagel-7b-v0.1}.\\

\concept{MetaMathDPO}
Paired version of the \texttt{MetaMathQA} dataset. To construct the paired preferences, the original responses are taken as the preferred completions and randomly corrupted (at an intermediate calculation) so that it is less preferable.

\subsubsection{MCQ Dataset}
We present \texttt{StemMCQ}, a modified version of the well-known \texttt{AQuA-RAT} dataset~\cite{ling2017program}, specifically designed to align the model with its primary purpose: answering STEM multiple-choice questions. The answers include the \texttt{AQuA-RAT} rationale followed by our extraction flag: \textit{"The answer is: <MCQ Letter>"}. We chose to include the rationale in our responses, as the Chain-of-Thought approach has demonstrated improved results compared to simply providing the answer~\cite{wei2022chain}. Table~\ref{tab:mcq_dataset} presents the dataset size.\\

\begin{table}[h]
    \centering
    \begin{tabular}{l c c}
        \toprule
        \textbf{Dataset} & \textbf{Size} & \textbf{Ratio} \\
        \midrule
        \texttt{AQuA-RAT} & 97,500 & 100\% \\
        \midrule
        \textbf{\texttt{StemMCQ} (ours)} & 97,500 & -- \\
        \bottomrule
    \end{tabular}
    \caption{Dataset sizes and their ratios of the MCQ dataset}
    \label{tab:mcq_dataset}
\end{table}

\concept{AQuA-RAT}
A large-scale dataset consisting of approximately 100,000 algebraic word problems. The solution to each question is explained step-by-step using natural language.

\subsubsection{DPR Dataset}
\label{sec:dpr_dataset}
To enable RAG in our model, we developed \texttt{StemDPR}, a DPR corpus of Wikipedia science documents. This dataset is built from \texttt{WikiStemCorpus}\footnote{See the dataset \href{https://www.kaggle.com/datasets/conjuring92/wiki-stem-corpus}{here}.}, a science-focused subset of the well-known RAG dataset \texttt{wiki\_dpr}~\cite{karpukhin-etal-2020-dense}. We compute the document embeddings of \texttt{WikiStemCorpus} using Facebook's DPR context encoder~\cite{karpukhin-etal-2020-dense}.

\subsection{Evaluation}
In this section, we define the evaluation process, which is divided into multiple steps. The initial step involves selecting the best model based on its generation quality. The final step assesses the predictability power of our MCQA model.

To select the best generation models, we need to assess the quality of their generation in terms of correctness and reasoning. 
\begin{itemize}
    \item \texttt{DPO Reward Accuracies} \cite{rafailov2023direct}: This allows us to assess the preference alignment of the model's generation in terms of human alignment.
\end{itemize}

To thoroughly assess our model's performance on \textit{STEM QA}, we choose diverse datasets that represent various skills the model should have acquired. First, we use benchmark datasets to evaluate the correctness of our first-stage model in answering open STEM questions:

\begin{itemize}
    \item \texttt{MATH} \cite{hendrycks2021measuring}: This dataset of 5k advanced mathematics questions to assess the model's mathematical step-by-step reasoning skills.
    \item \texttt{GSM8K} \cite{cobbe2021training}: A dataset of 8.5K (1k testing split) high quality linguistically diverse grade school math word problems created by human problem writers. Used to further evaluate the model's mathematical reasoning abilities.
\end{itemize}

Then, we use MCQA datasets to assess the MCQA performance of our final specialized model:
 
\begin{itemize}
    \item \texttt{MCQA Examples EPFL}: A dataset of around 350 samples designed to measure general knowledge and reasoning across multiple domains(It covers 57 subjects across STEM), used to test both world knowledge and problem solving ability.
\end{itemize}

We use accuracy as our metric since it was our target performance metric throughout the project and is best suited for evaluating unique MCQA answers.

\subsection{Baseline}
We compare \model{} with the candidate base models: \texttt{LLaMa-3-8B}~\cite{meta_llama3_2024} and \texttt{Mistral-7B}~\cite{jiang2023mistral}. At each step of the training pipeline (described in Section~\ref{sec:approach}), we conduct ablation studies by comparing the newly trained model with the model from the previous step.





\subsection{Setup}
We adapt our SFT and DPO procedures to run on a single V-100 GPU with 32 GB of VRAM and a single A-100 GPU with 40 GB of VRAM, respectively. We utilize the Unsloth library \cite{unslothai_unsloth_2024}, designed for fast and resource-efficient training of large language models. Combining Unsloth’s techniques with LoRa adaptors allows us to efficiently align LLaMa-3-8B and Mistral-7B within our resource constraints. In addition, due to the extended duration of the training process (more than 15 hours), extensive hyperparameter tuning is not practical. 

\subsubsection{1st SFT --  Mathematical Reasoning}
Therefore, the SFT hyperparameters (see Table~\ref{tab:sft_dpo_hyperparameters} in the appendix) are chosen based on the state-of-the-art SFT of the models. For similar reasons, we train our models using two relatively small, random sample sizes from the full SFT dataset (described in Section 2.1): 10,000 and 100,000 examples.

We conduct two SFT sessions for each model. The best models are selected from the 100,000-sample-size runs, showing the best results in the generation (see an example in \ref{appendix:gen}).

\subsubsection{DPO Alignment}
For the DPO training procedure, we split the DPO dataset described in Section \ref{sec:dpo_dataset} into 45,000 samples for training and the remaining for testing. The hyperparameter settings are described in Table \ref{tab:sft_dpo_hyperparameters}.

\subsubsection{2nd SFT --  MCQA Reasoning}
\begin{figure}[h!]
    \centering
        \includegraphics[width=0.45\textwidth]{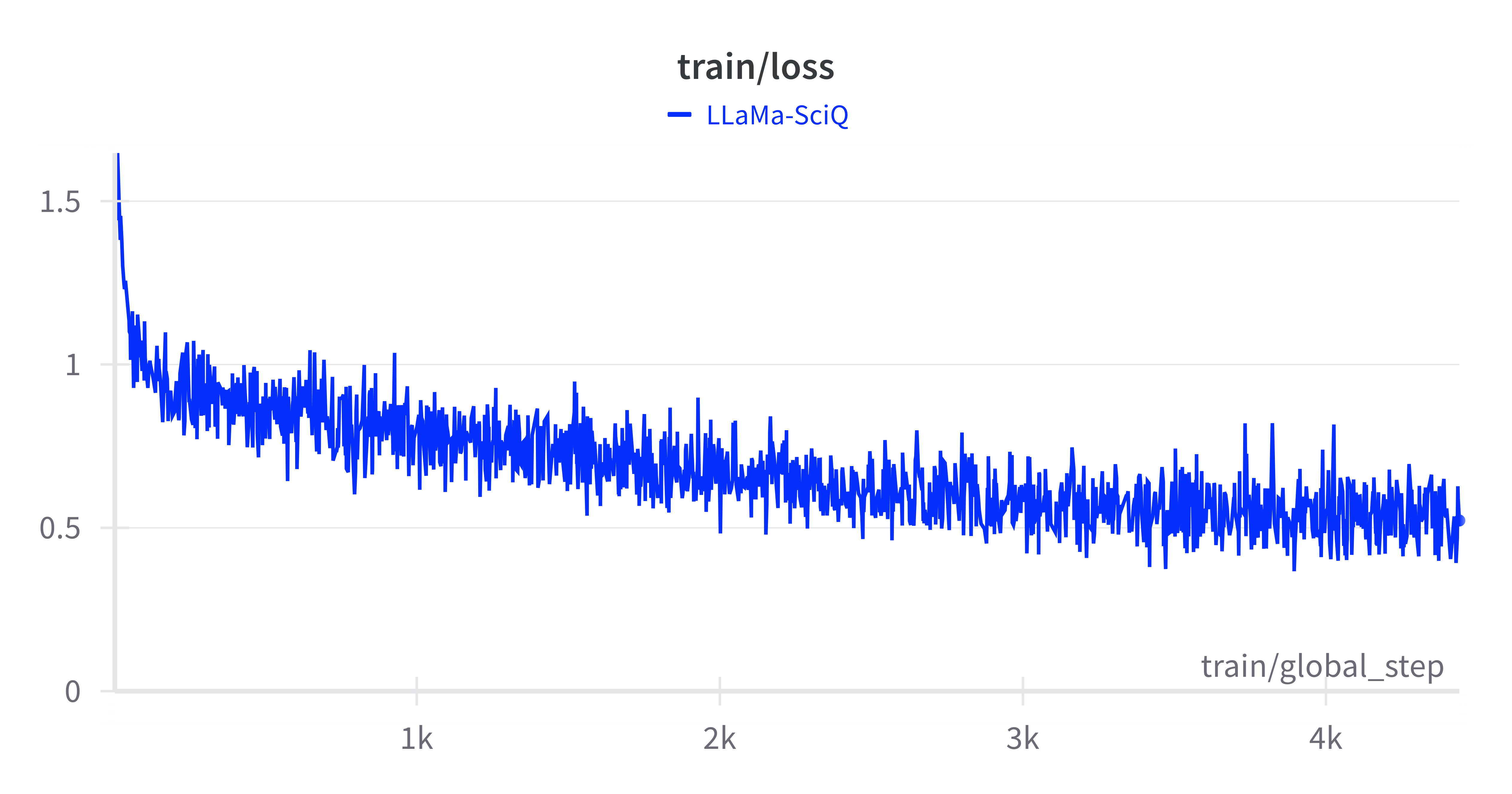}
        \caption{MCQ-SFT Training Loss}
        \label{fig:sft-mcq-loss}
\end{figure}

Finally, using the same hyperparameter setup as in the first SFT sessions, we perform the final SFT training for MCQA specialization using 97,500 MCQ samples. Figure~\ref{fig:sft-mcq-loss} presents the training loss of the kept run.

\subsection{Results}
The intermediate and final results can be found in Table~\ref{tab:results-mistral-llama-gsm8k}, Table~\ref{tab:results-mistral-llama-math}, and Table~\ref{tab:results-llama-epfl}.

\begin{itemize}
    \item \texttt{MATH} \cite{hendrycks2021measuring}: On the MATH dataset, known for its complexity and depth, we managed to achieve the performance announced by Meta on their introduction page of LLaMA-3-8B, with a score of 30\%. This demonstrates the power of LLaMA-3, especially in comparison to the Mistral-7B, where our results were consistent with Mistral's research, showing a score of around 11\%.
    \item \texttt{GSM8K} \cite{cobbe2021training}: For the GSM8K dataset, which is less challenging than MATH, our score was slightly below Meta's results by 5.1\%, but still more than 40\% higher than Mistral's performance. Note that we used 0-shot prompting for both evaluations, whereas Meta used few-shot prompting.
\end{itemize}

Finally, for the evaluation of \model{} on MCQA from the EPFL course, the results were decent but somewhat disappointing compared to the general math benchmarks.

\begin{itemize}
    \item \texttt{MCQA Examples EPFL}: The best score was achieved by the policy model, outperforming the two specializations by around 5\%. The RAG and Quantized models showed similar performance, with a difference of approximately 0.555 in accuracy. The RAG system did not improve accuracy and seemed to lead to poorer decisions, possibly due to the low similarity power of the retriever, inadequate content of the STEM DPR dataset, or the model's unfamiliarity with using context in prompts. However, the Quantized model, despite a significant reduction in size, only showed a 5\% loss in performance, which is a notable result.
\end{itemize}

\begin{table}[h!]
\centering
\begin{tabular}{ll}
\toprule
\textbf{Base Model} & \textbf{MATH}  \\
\midrule
LLaMa-3-8B-Instruct & \textbf{30\%} \textit{(4-shot, CoT)}\\
LLaMa-3-8B(SFT+DPO) & \textbf{30\%} \textit{(0-shot)}\\
Mistral-7B-Instruct & 11\% \textit{(4-shot, CoT)}\\
Mistral-7B(SFT+DPO) & 10.3\% \textit{(0-shot) }\\
\bottomrule
\end{tabular}
\caption{Performance comparison of different models on MATH benchmark}
\label{tab:results-mistral-llama-math}
\end{table}

\begin{table}[h!]
\centering
\begin{tabular}{ll}
\toprule
\textbf{Base Model} & \textbf{GSM8k} \\
\midrule
LLaMa-3-8B-Instruct & \textbf{79.6\%} \textit{(8-shot, CoT)}\\
LLaMa-3-8B(SFT+DPO) & 74.5\% \textit{(0-shot)}\\
Mistral-7B-Instruct & 39.9\% \textit{(8-shot, CoT)}\\
Mistral-7B(SFT+DPO) &  28.5\% \textit{(0-shot)}\\
\bottomrule
\end{tabular}
\caption{Performance comparison of different models on GSM8k benchmark}
\label{tab:results-mistral-llama-gsm8k}
\end{table}

\begin{table}[h!]
\centering
\begin{tabular}{ll}
\toprule
\textbf{Base Model} & \textbf{EPFL MCQA}  \\
\midrule
LLaMa-SciQ & \textbf{45.21\%} \textit{(0-shot)}\\
LLaMa-SciQ+RAG & 40.62\% \textit{(0-shot)}\\
LLaMa-Sci+Quanttize & 40.07\% \textit{(0-shot)}\\
\bottomrule
\end{tabular}
\caption{Models performance on MCQA EPFL benchmark}
\label{tab:results-llama-epfl}
\end{table}

\section{Analysis} 


We noted that our model exhibits reasonable generation capabilities and demonstrates sound reasoning when answering questions. During our SFT and DPO training, which frequently involved mathematical questions, our model proved particularly adept at handling them. However, as the benchmark (Table \ref{tab:results-llama-epfl}) included questions from a wide range of disciplines, the results were generally acceptable.

We believe our quantized model maintained the accuracy of the LLaMa-SciQ model, as it occasionally achieved higher accuracy in our tests. During development, we experimented with various configuration settings, including adjustments to the temperature, to optimize performance. Table~\ref{tab:gen_config_small_res} presents the best results of the generation tested on a 10-subsample of the EPFL MCQ dataset; the full test results are presented in Appendix~\ref{appendix:generation_tune}). Despite these efforts, we think the generation configuration could still benefit from fine-tuning. With a large beam size in the beam search, the quantized model's performance was comparable to that of LLaMa-SciQ. However, due to resource constraints, we reduced the beam size to 1 for our final benchmark.

The RAG model did not meet our goal of enhancing accuracy. We attribute this to the encoder used for information retrieval, which was not specifically fine-tuned for our model. Consequently, the encoder sometimes retrieved irrelevant information, potentially biasing the model towards incorrect data. Additionally, our model was trained to adhere to a specific template rather than to utilize the provided information effectively, which likely contributed to its underperformance.

\begin{table}[h]
    \centering
    \begin{tabular}{p{4.5cm} p{2.5cm}}
        \toprule
        \textbf{Generation Configuration} & \textbf{Accuracy} \\ 
        \midrule
        Greedy & 40\% \\ 
        Sample (default) & 40\% \\ 
        \textbf{Sample (default, temp=0.3)} & \textbf{50\%} \\ 
        Sample (default, top\_p=0.95, temp=0.3) & 40\% \\ 
        \bottomrule
    \end{tabular}
    \caption{Accuracy of Different Sampling Methods on the 10-sample of EFPL MCQA}
    \label{tab:gen_config_small_res}
\end{table}

\section{Ethical considerations}
\label{sec:ethics}
In this section, we address the ethical considerations relevant to \model{}.\\

\concept{Low-Resources Language Performances}The high performance of LLaMa-3-8b on high-resource languages~\cite{meta_llama3_2024} suggests that \model{} should be capable of handling questions in most of these languages (with the best performance on English MCQs, as the SFT dataset is English-based). However, additional work is needed to extend its capabilities to low-resource languages, such as Urdu and Swahili. This could be achieved by expanding our SFT datasets to teach the model multilingual scientific reasoning. Furthermore, although more challenging and costly, we could extend our DPO dataset to include low-resource languages preferences to improve the model's generations in these latest.\\

\concept{Accessibility for Deaf Community} The exclusion of signed languages from modern language technologies marginalizes Deaf communities, who prefer to communicate in signed languages online~\cite{yin2021including}. Therefore, it is essential to include signed language compatibility in our model to respect this community and support its communication preferences. One potential approach to achieve this is by harnessing Sign Language Translation (SLT), which has seen advancements through deep learning techniques~\cite{al2021deep, chen2022simple}, such as the \texttt{STMC-Transformer} model~\cite{yin2020better}. By integrating SLT into \model{}'s pipeline, we could easily address signed questions. \\

\concept{Social Bias \& Harmful Content} The model, designed for the MCQA task, should not exhibit more harmful content or social bias than its inherent base model. However, for broader usages, studies indicated that LLM presents vulnerabilities exploitable to output harmful content or social bias~\cite{wei2024jailbroken, deng2023jailbreaker}. Therefore, future work should involve additional training to mitigate \model{}'s potential biases or harmful content that may arise from out-of-scope usages. This can be achieved using Meta's Responsible Use Guide (RUG)\footnote{See the RUG \href{https://llama.meta.com/responsible-use-guide/}{here}.} and LLaMa-Guard~\cite{inan2023llama}, an LLM-based safeguard model designed for Human-AI conversation use cases.

\section{Conclusion}
\label{sec:conclusion}
In this work, we propose \model{}: an educational chatbot designed for science multiple-choice question answering (MCQA). The model is a fine-tuned \texttt{LLaMa-3-8B} aligned with human preferences using the novel STEM datasets introduced (\texttt{StemQA}, \texttt{StemDPO}, \texttt{StemMCQ}). It also employs cost-reducing training techniques such as Unsloth~\cite{unslothai_unsloth_2024} to address limitation in resources. \model{} maintains the performance of state-of-the-art large language models in scientific question answering, achieving up to 74.5\% on the GSM8k benchmark and 30\% on the MATH benchmark using zero-shot prompting. These results are comparable to the base model using eight-shot prompting on these benchmarks. Exploring few-shot prompting could be a promising direction for future work. While the model's performance on the MCQA task yielded relatively low results, they are acceptable considering the complexity of such a specialized task.

Future work includes enhancing the model's performance by exploring various prompting strategies~\cite{wang2022self, wan2023better}. Additionally, adapting the model to more languages -- with an emphasis on signed languages -- and evaluating its social biases will be essential to make it accessible to all, thereby strengthening its educational impact.
\clearpage

\bibliography{anthology,custom}
\bibliographystyle{acl_natbib}

\newpage
\appendix
\input{appendix.tex}

\end{document}

%% file: appendix.tex
\section{Contribution}
Each member of the group contributed equally to all of the aspects of the project.\\
Matin Ansaripour: DPO training, LLaMa-adapter supervised training, Quantization coding, Report Writing.\\
Paul Teiletche: Dataset processing, external dataset adaptation, RAG specialisation, Report writing, Evaluation coding.\\
Marc-Antoine Allard: Dataset processing, external dataset adaptation, RAG specialisation, Report writing, Evaluation coding.\\
Maria Yuffa: DPO training, Quantization coding, Literature review, Report writing.

\section{Datasets Samples}
\label{appendix:dataset_examples}
\subsection{SFT Dataset}
\sample{
    \texttt{
        \textbf{"problem"}: "Determine the sum of the positive factors of 48.",\\
        \textbf{"solution"}: "To find the sum of the positive factors of 48, we can [...]. The answer is: 124"
    }
}{SFT dataset}
\subsection{DPO Dataset}
\sample{
    \texttt{
        \textbf{"prompt"}: "Tom eats a pound of carrots [...] how many calories did he eat in total?",\\
        \textbf{"chosen"}: "Tom eats 1 pound of carrots, which have 51 calories per pound, so he eats 1*51 = 51 calories [...] The answer is: 85",\\
        \textbf{"rejected"}: "Tom eats 1 pound of carrots, which have 51 calories per pound, so he eats 1*51 = 97 calories [...] The answer is: 85"
    }
}{DPO dataset}
\subsection{MCQ Dataset}
\sample{
    \texttt{
        \textbf{"subject"}: "maths",\\
        \textbf{"question"}: "There are 8 players in a chess group [...] how many total games will be played?",\\
        \textbf{"options"}: ["10","30","28","60","90"]\\
        \textbf{"answer"}: "10 players are there. two players [...] The answer is: C."
    }
}{MCQ dataset}

\subsection{DPR Dataset}
\sample{
    \texttt{
        \textbf{"text"}: "In mathematical analysis, the Cauchy index is [...] the degree of q.",\\
        \textbf{"title"}: "Cauchy index"\\
        \textbf{"embeddings"}: [-0.6179105639457703, ..., 0.35533231496810913]
    }
}{DPR dataset}

\subsection{Intruction for DPO Generation}
\label{appendix:dpo_examples_instruct}
\textbox{"Imagine you're a teaching assistant for a \textit{<course\_topic>} course. A student has just asked the question above. Your goal is to provide a comprehensive and detailed explanation, similar to how you would guide a student in understanding the concept thoroughly. Use scientific reasoning and relevant examples to clarify the topic and ensure a deep understanding by the student."}{Instruction to generate examples for DPO}

\section{Training Details}
Here we present more details for SFT and DPO training. 

\subsection{Training Hyperparameters}
\label{appendix:train_metrics}
Table \ref{tab:sft_dpo_hyperparameters} presents the hyperparameters that we used for each training.
\begin{table}[h!]
\centering
\begin{tabular}{lll}
\toprule
\textbf{Hyperparameter} & \textbf{SFT Values} & \textbf{DPO Values} \\
\midrule
Epochs & 1 & 1 \\
Batch Size & 4 & 2 \\
Warmup Ratio & 0.1 & 0.1 \\
Learning Rate & 2e-4 & 5e-5 \\
LR Scheduler & Linear & Cosine \\
Weight Decay & 1e-2 & 1e-2 \\
Neftune Noise $\alpha$ & 5 & - \\
GA Steps & 1 & 4 \\
\bottomrule
\end{tabular}
\caption{SFT and DPO Hyperparameters}
\label{tab:sft_dpo_hyperparameters}
\end{table}

\subsection{Training Metrics}
\subsubsection{Maths-SFT}
\begin{figure}[h!]
    \centering
    \begin{subfigure}[b]{0.45\textwidth}
        \includegraphics[width=\textwidth]{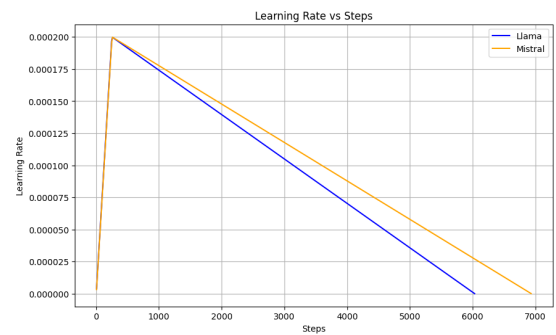}
        \caption{Learning Rate vs Steps}
        \label{fig:lr_sft}
    \end{subfigure}
    \hfill
    \begin{subfigure}[b]{0.45\textwidth}
        \includegraphics[width=\textwidth]{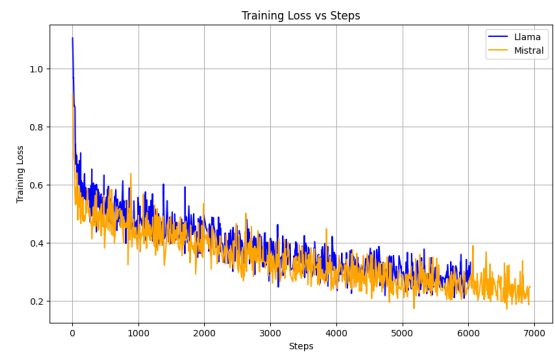}
        \caption{Training Loss vs Steps}
        \label{fig:train_loss_sft}
    \end{subfigure}
    \caption{SFT Training statistics for Llama and Mistral models on 100,000 samples.}
    \label{fig:sft_metrics}
\end{figure}

Figure~\ref{fig:sft_metrics} presents the most important training metrics values of the best Maths-SFT runs of each model.
\subsubsection{MCQ-SFT}
\begin{figure}[h]
    \centering
    \begin{subfigure}[b]{\hsize}
        \includegraphics[width=\textwidth]{image/llamasciq-loss.png}
        \caption{Training Loss}
    \end{subfigure}
    \hfill
    \begin{subfigure}[b]{\hsize}
        \includegraphics[width=\textwidth]{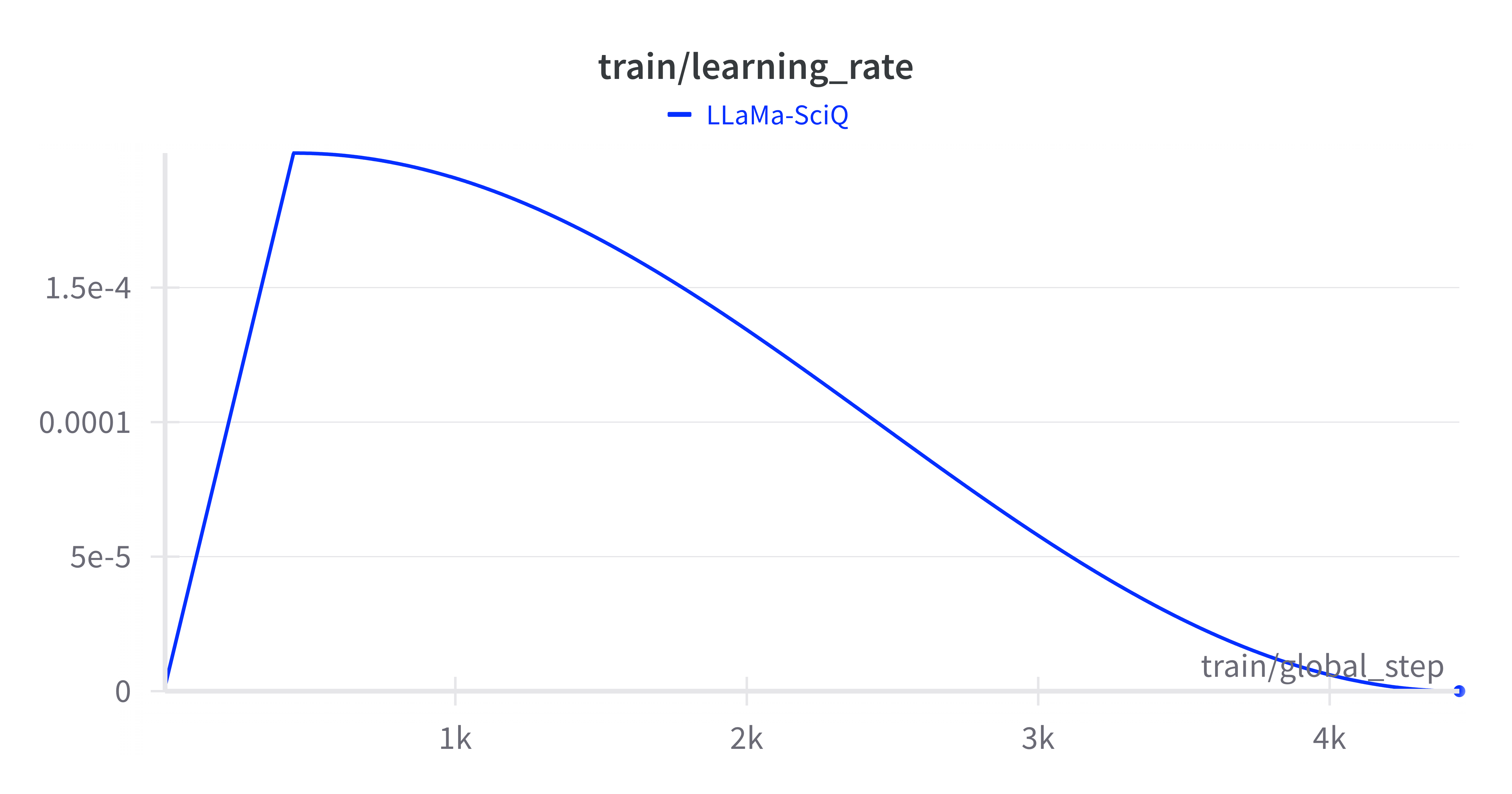}
        \caption{Learning Rate}
    \end{subfigure}
    \begin{subfigure}[b]{\hsize}
        \includegraphics[width=\textwidth]{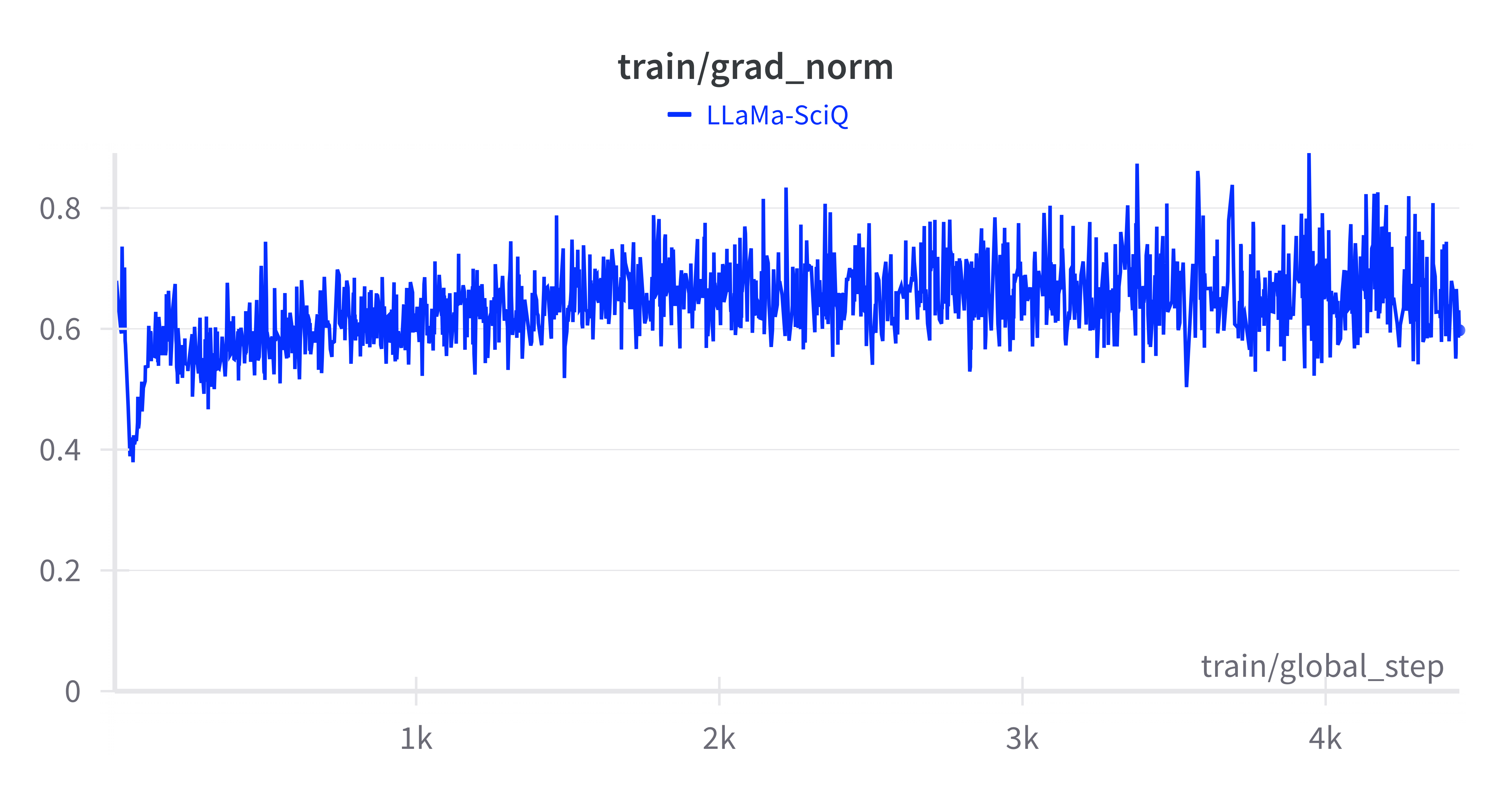}
        \caption{Gradients Norm}
    \end{subfigure}
    \caption{Training Analytics: Transformers Models}
    \label{fig:sft_analytics}
\end{figure}

Figure~\ref{fig:sft_analytics} presents the training metrics of the MCQ-SFT.

\section{Model Samples}
\label{appendix:prompt}

\subsection{Prompt Format}
\label{appendix:prompt}
Below is the prompt format used for the generations. If a section is not relevant for the task (e.g. context for non-RAG generation or options for non-MCQ), it is removed.

\textbox{
    \textbf{\texttt{Context Information:}}\\
    \{RAG Context\}\\
    
    Below is a MCQ that you will need to answer using the above context information. Write an answer that fully explains your reasoning.\\
    
    \textbf{\texttt{\#\#\# Question:}}\\
    \{Question\}\\
    
    \textbf{\texttt{\#\#\# Options:}}\\
    \{Options\}\\
    
    \textbf{\texttt{\#\#\# Answer:}}\\
    \{Answer\}}{Prompt Format}

\subsection{Generation}
\label{appendix:gen}

Here we present a sample of our LLaMa-3 maths fine-tuned model's generation. The blue part is the one generated by the model.
\generation{
    Below is a question that you will need to answer. Write an answer that fully explains your reasoning.
    }{
    The town of Belize has 400 homes. One fourth of the town's homes are white. One fifth of the non-white homes have a fireplace. How many of the non-white homes do not have a fireplace?
    }{
    \textcolor{blue}{One fourth of the town's homes are white, so there are 400/4 = 100 white homes. The remaining homes are non-white, so there are 400 - 100 = 300 non-white homes. One fifth of the non-white homes have a fireplace, so there are 300/5 = 60 non-white homes with a fireplace. Therefore, the number of non-white homes that do not have a fireplace is 300 - 60 = 240. \#\#\# 240 The answer is: 240} 
    }{Aligned LLaMa-3 Model Generation}

\section{Generation Tuning}
\label{appendix:generation_tune}
Table~\ref{tab:gen_tune_res} presents the MCQA accuracy of each configuration tested on a 10-subset of the EPFL MCQA dataset
\begin{table}[h]
    \centering
    \begin{tabular}{p{4.5cm} p{2.5cm}}
        \toprule
        \textbf{Generation Configuration} & \textbf{Accuracy} \\ 
        \midrule
        Greedy & 40\% \\ 
        Beam Search (B=3) & 30\% \\ 
        Beam Search (B=5) & 30\% \\ 
        Sample (default) & 40\% \\ 
        \textbf{Sample (default, temp=0.3)} & \textbf{50\%} \\ 
        Sample (default, temp=0.1) & 40\% \\ 
        Sample (default, temp=1.2) & 30\% \\ 
        Sample (default, top\_p=0.95, temp=0.3) & 40\% \\ 
        Sample (default, top\_p=0.85, temp=0.3) & 40\% \\ 
        \bottomrule
    \end{tabular}
    \caption{Accuracy of Different Sampling Methods on 10-sample of EFPL MCQA}
    \label{tab:gen_tune_res}
\end{table}